\documentclass[letterpaper, 10 pt, conference]{ieeeconf}  

\usepackage{graphicx}
\usepackage{cite}
\usepackage{amsmath}
\usepackage{multirow}
\DeclareMathOperator*{\argmin}{arg\ min}

\IEEEoverridecommandlockouts         

\title{\LARGE \bf
Towards Better Performance and More Explainable Uncertainty for 3D Object Detection of Autonomous Vehicles}

\author{Hujie Pan$^{1, 2}$, Zining Wang$^{2}$, Wei Zhan$^{2}$, and Masayoshi Tomizuka$^{2}$
\thanks{$^{1}$Hujie Pan is with School of Mechanical Engineering, Shanghai Jiao Tong University, Shanghai, 200240, China and now a visiting student researcher in University of California Berkeley, Berkeley, 94720, USA
        {\tt\small phj1991@sjtu.edu.cn}, {\tt\small phoenvujih@berkeley.edu} }%
\thanks{$^{2}$Zining Wang, Wei Zhan and Masayoshi Tomizuka are with Department of Mechanical Engineering, University of California Berkeley, Berkeley, 94720, USA 
        {\tt\small wangzining@berkeley.edu}, {\tt\small wzhan@berkeley.edu}, {\tt\small tomizuka@berkeley.edu}}%
}

\begin{document}

\maketitle
\thispagestyle{empty}
\pagestyle{empty}

\begin{abstract}

In this paper, we propose a novel form of the loss function to increase the performance of LiDAR-based 3D object detection and obtain more explainable and convincing uncertainty for the prediction. The loss function was designed using corner transformation and uncertainty modeling. With the new loss function, the performance of our method on the \textit{val} split of KITTI dataset shows up to a 15\% increase in terms of Average Precision (AP) comparing with the baseline using simple L1 Loss. In the study of the characteristics of predicted uncertainties, we find that generally more accurate prediction of the bounding box is accompanied by lower uncertainty. The distribution of corner uncertainties agrees on the distribution of the point cloud in the bounding box, which means the corner with denser observed points has lower uncertainty. Moreover, our method learns the constraint from the cuboid geometry of the bounding box in the uncertainty prediction. Finally, we propose an efficient Bayesian updating method to recover the uncertainty for the original parameters of the bounding boxes which can help provide probabilistic results for the tracking and planning module.

\end{abstract}

\section{INTRODUCTION}

The “perception-plan-control” scheme has been widely adopted as the framework for autonomous driving solutions \cite{pendleton2017perception}. As a significant step in the scheme, the perception module provides an understanding of the environment within which the object detection and localization play crucial roles. With the help of onboard sensors such as camera and LiDAR, the color and depth information of the foreground and background can be obtained. A lot of efforts have been made to the object detection algorithms using the sensor data above, among which deep learning methods have made great progress in precision and callback rate \cite{chen2017multi,feng2020deep,meyer2019lasernet,qi2018frustum,shi2020pv,shi2019pointrcnn,zhou2018voxelnet}. Considering the variation of the environment, states of the objects and different observability, the predicted result should include a certain level of uncertainty, which is also an essential input for the planning and decision-making modules. However, most deep learning based detectors only produced the deterministic states of the object while lacked feedback of uncertainties \cite{schwarting2018planning}. 

To tackle the problem mentioned above, a lot of attempts have been made to quantitatively predict the uncertainty for Deep Neural Networks (DNN) and some of them are proposed to predict the uncertainties of different parameters for object detection. \cite{meyer2019learning,he2019bounding} proposed a method to learn the probabilistic distribution of the parameters by minimizing the Kullback-Leibler divergence (KLD) of the predicted distributions from preset ones. Method \cite{meyer2019learning} considers the variations of labeled point data while highly depends on how the probabilistic distribution is preset. \cite{wang2020inferring} infers the uncertainty of the label based on the point cloud distribution of the sample using Bayesian method and produces a probabilistic map to represent the uncertainty of the bounding box. \cite{meyer2019lasernet,feng2018towards,feng2019leveraging} directly learn the parameters of the probability distribution by maximizing the likelihood. However, they all assume that the parameters are independent with each other and adopt a diagonal covariance matrix for uncertainty modeling. \cite{feng2019leveraging} further analyzes the features of the predicted uncertainties but is constrained by the independence assumption. This assumption makes the derived ensemble variance of each corner for the bounding box all the same. In that case, the modeled uncertainty is unable to fully reflect the distribution of the point cloud which makes it less explainable and persuasive. A non-diagonal assumption could potentially address this issue but is prone to numerical instability such as gradient exploding in our preliminary test. 

Instead of the non-diagonal covariance assumption for the original parameters of the bounding box, we propose a method that first transfer the original parameters to the eight corners and model the probabilistic distribution of the location of each corner. This method provided enough degrees of freedom (DOFs) in representing the uncertainties and avoid the numerical instability of training a non-diagonal covariance matrix. As for the network architecture, following PointRCNN, we proposed a PointNet-based 2-stage method which keeps lossless point-wise features. 

With the proposed approach, we are able to considerably improve the performance of the object detector from the baseline and reach a comparable average precision (AP) with the state-of-the-art algorithms. Meanwhile, the predicted uncertainty successfully represented the distribution of the points in the bounding box as well as the constraint of its cuboid geometry. Finally, we proposed a Bayesian updating method to recover the uncertainty of the original parameter set of the bounding box so as to provide the uncertainty of states for tracking and planning.

\section{RELATED WORKS}

\subsection{LiDAR-based 3D object detection}

Unlike the organized pixel values in images, point cloud provides irregular data which prevents the direct application of classical convolutional neural networks (CNN) such as VGG, ResNet. To tackle this issue, many researchers preprocess the point cloud data by reorganizing it into 2D or 3D grids \cite{chen2017multi, zhou2018voxelnet, lang2019pointpillars, yan2018second, yang2018pixor}. However, voxelization resulted in large size of the input and high computation cost until the 3D sparse convolution method was proposed \cite{yan2018second, graham20183d}. Moreover, the grid-based methods potentially cause information loss of point cloud and have limited receptive field. To make use of lossless point cloud information, PointNet-based architectures are proposed to extract point-wise and global features from the point cloud \cite{qi2018frustum, qi2017pointnet, qi2017pointnet++}. There are several 2-stage object detectors based on PointNet such as PointRCNN \cite{shi2019pointrcnn} and STD \cite{yang2019std}. In this work, we propose a 2-stage method that utilize point-wise features for region proposal and refinement following PointRCNN. 

\subsection{Uncertainty modeling for deep learning}

To represent the uncertainty of neural networks, Bayesian modeling is proposed to estimate the epistemic and aleatoric uncertainties \cite{mackay1992practical,kendall2015bayesian, kendall2017uncertainties}. Epistemic uncertainty is model-based which arises from the uncertainty of model parameters. It reflects the limitation of the model on describing the biased training data and can be reduced by enlarging the dataset \cite{kendall2017uncertainties}. There are two main ways to predict the epistemic uncertainty: variational inference \cite{graves2011practical} and sampling \cite{gal2016dropout}. The aleatoric uncertainty, on the other hand, is measurement-based which arises from the sensor noise, data representation and label noise, etc. \cite{malinin2018predictive}. It can be predicted by outputting the parameters of a distribution and the method is adopted in this paper. 
Recently, many efforts have been put on estimating the uncertainty of 3D object detection. \cite{feng2018towards} and \cite{miller2018dropout} utilized Monte Carlo dropout to capture the epistemic uncertainty of the bounding box. While \cite{meyer2019lasernet} and \cite{feng2019leveraging} predicted the aleatoric uncertainty by learning the parameters in the probabilistic distribution of bounding boxes. \cite{meyer2019learning} and \cite{he2019bounding} further pre-estimated the uncertainty of the label and learned the probabilistic distribution by minimizing the KLD of predicted from preset ones. In this work, we estimate the aleatoric uncertainties of the corners from bounding boxes by learning the parameters of the probabilistic distribution and analyze how they represent the distributions of point clouds and are constrained by the cuboid geometry. Finally, we propose a Bayesian update method to recover the uncertainty of the original parameters in labels.

\section{Proposed Method}
In this section, we propose a two-stage detector whose RPN stage and the point cloud encoder use the backbone PointRCNN \cite{shi2019pointrcnn} and PointNet++ \cite{qi2017pointnet++}. It is trained by the proposed innovative form of output and loss function with corner transformation and uncertainty modeling. The architecture of the network is shown in Fig. \ref{network}. 

\subsection{Network Architecture}
\begin{figure}[thpb]
  \centering
  \includegraphics[scale=1.0]{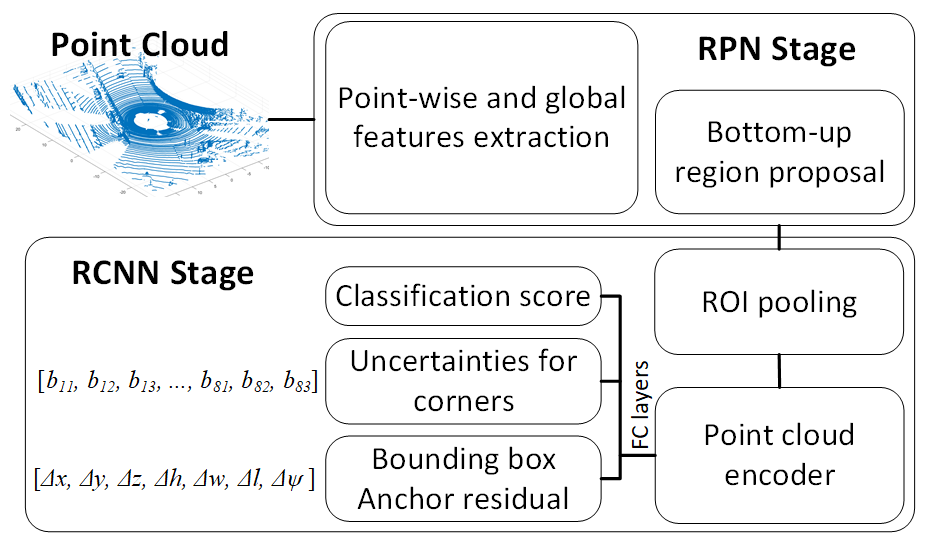}
  \caption{The architecture of the network that predicts the uncertainty of the corners of 3D bounding boxes and the residual of bounding boxes.}
  \label{network}
\end{figure}

As shown in Fig. \ref{network}, we take raw point cloud data as input rather than voxelizing it. In the region proposal network (RPN) stage, we generate 3D regions of interest (ROIs) using the point-wise and global features extracted by a PointNet-like architecture. After ROI-pooling, we feed the pooled point features along with the original point coordinates and intensities to the point cloud encoder of PointNet++ \cite{qi2017pointnet++} to learn the representation of the point cloud. The encoder is followed by three task-specified fully connected (FC) layers which output the classification scores, box residuals and the uncertainties of the locations of box corners, respectively. 
The FC layers only predict the residual of the bounding box. The final bounding box is recovered by combining the residual and the preset anchor of the ROI pooling layer together.

The labels of KITTI dataset describe the ground truth bounding box with 7 parameters: 3 for the center location, 3 for the dimensions and 1 for the orientation. The orientation is described by the yaw angle of the object. Accordingly, our ROI box and final refined bounding box are all parametrized by $[x, y, z, h, w, l, \psi]$ in which $[x, y, z]$ denotes the location, $[h,w,l]$  denotes the dimensions and $\psi$  is the yaw angle. 

\subsection{Corner Transformation}

A corner-based equally weighted regression loss is proposed to enrich the representation capability of uncertainties. Rather than directly predicting the coordinate of each corner, we keep the original expression of the bounding box and transformed it to the corner coordinates in camera coordinate system while retaining the cuboid constraints of them. 

Equation \ref{corntrans} is the transforming function from the original label parameters to the corresponding 8 corners.
\begin{equation}
\begin{bmatrix}
x_c\\ y_c\\ z_c
\end{bmatrix}
=\begin{bmatrix}
\sin \psi &0  & \cos \psi \\ 0 & 1 & 0 \\ -\cos \psi & 0 & \sin \psi 
\end{bmatrix}
\begin{bmatrix}
\pm l/2\\ \pm h/2\\ \pm w/2
\end{bmatrix}
+\begin{bmatrix}
x\\ y\\ z
\end{bmatrix} 
\label{corntrans}
\end{equation}
in which $ [x_c, y_c, z_c]^{T} $   is the location of the corner, $l, h, w$ are the dimensions of the bounding box, $ \psi$ is the yaw angle and $ [x, y, z]^T $ is the location of the box center. We transform both the label and recovered bounding box for loss calculation. 

\subsection{Uncertainty Modeling}

Denote the procedure from sensing to annotation as a measurement, and the predicted bounding box as the mean value, we assume the components of the coordinates of corners to be drawn from independent univariate Laplace distributions with a probability density function defined as follows: 
\begin{equation}
    p(x|\mu, b)=\frac{1}{2b} \exp{(-\frac{|x-\mu|}{b})}
    \label{laplace}
\end{equation}
in which $\mu$ is the predicted transformed component, $b$ is the predicted diversity of the Laplace distribution. The variance of the Laplace distribution equals to $2b^2$. 

Then we take the negative log likelihood as the loss function for each single component of the corner: 
\begin{equation}
\mathcal{L} (x,\mu, b) = -\log{p(x|\mu,b)} = \ln{2b} +\frac{|x-\mu|}{b}
\end{equation}

Finally, we calculate the ensemble regression loss of the 3D bounding box by summing up the loss of all the components from all corners:
\begin{equation}
    \mathcal{L} _{ens} = \sum_{i}\sum_{j} \mathcal{L}(x_{ij},\mu_{ij},b_{ij})
\end{equation}
in which $i$ is the index of the corner, and $j$ is the index of the component of the corner. 

Since the form of loss is negative log likelihood, summing up the losses is equivalent to multiplying the probability density defined in (\ref{laplace}). In other words, by minimizing the ensemble loss, we are maximizing the likelihood of labeled corners under the Laplace distribution with parameters $\mu$, $b$.  

\section{EXPERIMENTAL RESULTS}
This section starts with the evaluation of our method on 3D object detection benchmark of the ‘Car’ category on KITTI dataset \cite{geiger2012we}. The dataset provides 7481 training samples and 7518 testing samples. Following \cite{shi2019pointrcnn}, we divide the training data into \textit{train} split (3712 samples) and \textit{val} split (3769 samples). We conduct an ablation study with different cases on \textit{val} split and compare our method with the state-of-the-art algorithms on \textit{test} set. After the evaluation, we further analyze the behavior of the uncertainties as well as how it represents the distribution of the point cloud and how it is constrained by cuboid geometry. 

\subsection{3D Object Detection on KITTI}
We first compare the performance of our method with the state-of-the-art methods on KITTI \textit{test} set. We pick some representative LiDAR-based methods listed in Tab. \ref{test}. We highlight the comparison of performance between our method and PointRCNN which uses bin-based loss function and serves as the base for our method. 

All results are evaluated by the average precision (AP) with a 3D intersection over union (IoU) threshold of 0.7 on easy, moderate, and hard difficulty levels respectively. The AP of our method on \textit{test} set was calculated on official KITTI server.

As shown in Tab. \ref{test}, our method achieves a comparable level of performance with the methods listed and surpasses most of them. When compared with the base network PointRCNN, our method has a comparable performance with it at easy difficulty level. With the increase of the difficulty, our method surpasses PointRCNN in terms of AP by 1.23\% at moderate and 2.47\% at hard difficulty levels which indicates not only a better performance but also higher robustness for less informative samples.

To find out how corner transformation and uncertainty modeling affect the performance of the model, we perform an ablation study in different cases on the \textit{val} split. As shown in Tab. \ref{ablation}, we set a baseline whose loss function was simply L1 form without corner transformation and uncertainty modeling. The baseline with corner transformation is proceeded with the L1 loss calculation on the components of transformed corners. The baseline with uncertainty adopts only the aleatoric uncertainty modeling without any other modifications following \cite{feng2019leveraging}. To make a fair comparison, the results of different cases share the same RPN result and follow the same training procedure. We also add the performance of PointRCNN on \textit{val} split coming from \cite{shi2019pointrcnn} for comparison. 

\begin{table}[h]
\begin{center}
\caption{Validation results on the KITTI test dataset}
\label{test}
\begin{tabular}{c|ccc} 
\hline
\multirow{2}{*}{\textbf{Method}} & \multicolumn{3}{c}{\textbf{Average Precision 3D (\%)}}  \\
& \textbf{\textit{Easy}}{\centering} & \textbf{\textit{Moderate}} & \textbf{\textit{Hard}}   \\ 
\hline
  MV3D \cite{chen2017multi}                &74.97	&63.63	&54.00    \\
  SECOND \cite{yan2018second}              &83.34	&72.55	&65.82    \\
  PointPillars \cite{lang2019pointpillars} &82.58	&74.31	&68.99    \\
  STD \cite{yang2019std}                   &87.95	&79.71	&75.09     \\ 
\hline
  PointRCNN \cite{shi2019pointrcnn}        &\textbf{86.96}	&75.64	&70.70      \\
  \textbf{Ours}                            &86.55	&76.87 (\textbf{+1.23})	&73.17 (\textbf{+2.47})      \\
\hline
\end{tabular}
\end{center}
\end{table}

\begin{table}[h]
\begin{center}
\caption{Ablation study on the \textit{VAL} split}
\label{ablation}
\begin{tabular}{c|ccc} 
\hline
\multirow{2}{*}{\textbf{Method}} & \multicolumn{3}{c}{\textbf{Average Precision 3D (\%)}}  \\
& \textbf{\textit{Easy}}{\centering} & \textbf{\textit{Moderate}} & \textbf{\textit{Hard}}   \\ 
\hline
  Simpe L1 loss (\textbf{baseline})      &73.03	&66.02	&60.86    \\
  Baseline with \textbf{corner trans}    &85.74	&77.64	&76.05    \\
  Baseline with \textbf{uncertainty}     &83.85	&77.47	&74.57    \\
\hline
  PointRCNN (\textbf{bin-based loss}) \cite{shi2019pointrcnn}  &88.88	&78.63	&77.38      \\
  \textbf{Corner trans \& uncertainty (ours)} &\textbf{89.26}	&\textbf{80.62} &\textbf{79.13}     \\
\hline
\end{tabular}
\end{center}
\end{table}

As shown in Tab. \ref{ablation}, corner transformation and uncertainty modeling improve the performance of the model by 11.62\% and 11.45\% respectively on moderate difficulty level from the baseline. While our method with both the above features improves the AP by 16.23\%, 14.60\% and 18.27\% on easy, moderate and hard difficulty levels respectively from the baseline. 
The corner transformation increases the performance of the model by re-weighting the original parameters of the bounding boxes and transforming them into equally weighted corners. That’s because the corners contribute equally in representing the bounding box such as IoU calculation, while the original parameters are not. In that case, even with simple L1 loss function, corner transformation considerably increases the performance of the model from the baseline. On the other hand, uncertainty modeling helps to increase the noise tolerance of the detector which means the noisy label affected less on updating the network weights due to the diversity of Laplace distribution. Combining these two methods, our method increases the performance by 0.4\%, 1.99\% and 1.75\% on easy, moderate and hard difficulty levels respectively from PointRCNN with bin-based loss on \textit{val} split. 

\subsection{Explaining the Uncertainties}

In this section, we analyze the characteristics of the predicted uncertainty starting with its general behaviors. Then, we discuss the relationship between the corner uncertainty and the point cloud distribution. Finally, we introduce how cuboid constraint affects the distribution of the corner uncertainties.

\subsubsection{General Behaviors}

\begin{figure}[thpb]
  \centering
  \includegraphics[width=8cm]{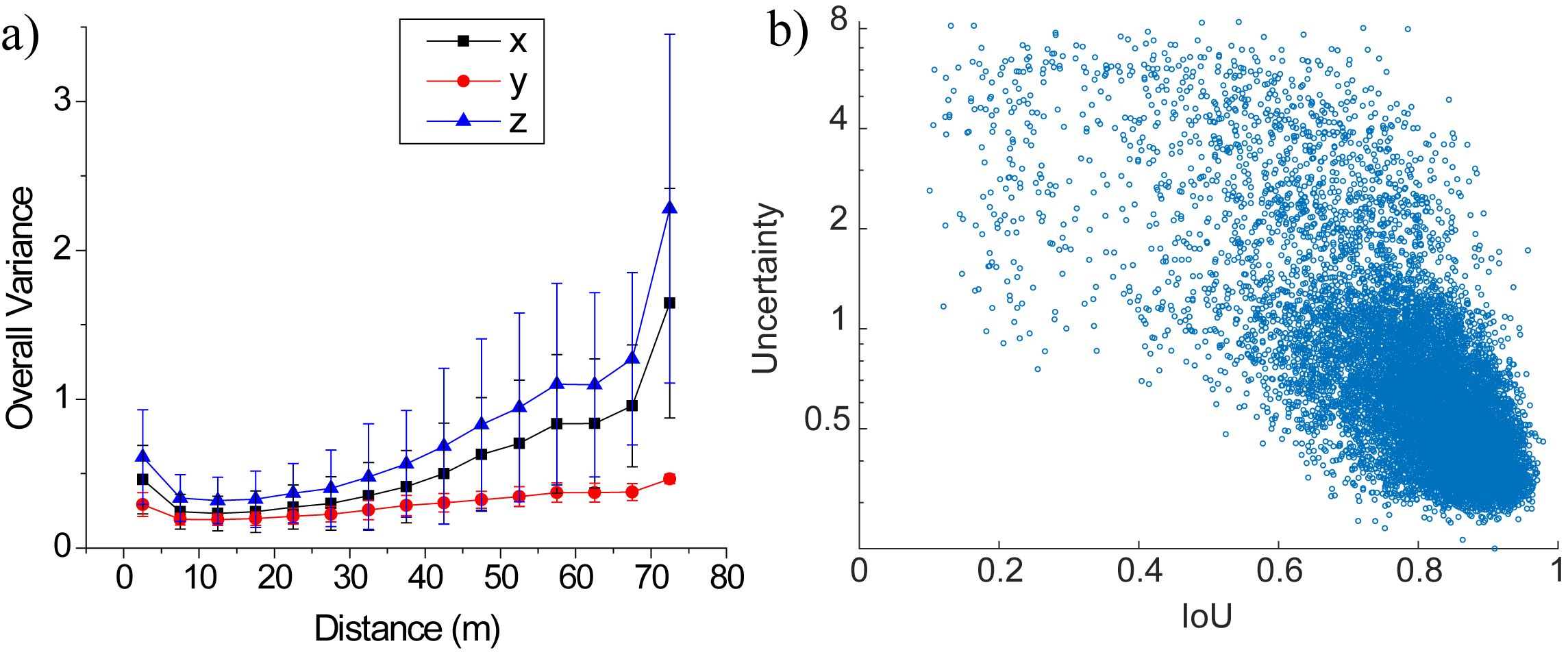}
  \caption{The general behavior of the uncertainty: a) the overall variance of each component from the corners with respect to the detection distance, b) the total uncertainty with respect to IoU of the bounding boxes. The total uncertainty of the bounding box is calculated by summing up all the corner variances}
  \label{fig2}
\end{figure}

To reveal the relationship between the distance to ego-vehicle and the uncertainty of different components, we calculate the overall variance of each bounding box by summing up the variances of a single component from its eight corners. Notice that here the $x$, $y$ and $z$ components are obtained in the camera coordinate system. As shown in Fig. \ref{fig2}, we plot the mean values of overall variance in bins for every 5 meters with respect to the distance. It shows that for corners with distances less than 10 meters, the overall variance decreases as the distance increases. It is attributed to the truncated objects close to the LiDAR. While the distance is greater than 10 meters, the variances of all 3 components increase with the increase of the distance. Moreover, we calculate the standard deviation of the overall variances to represent its variation and plot it as error bars in Fig. \ref{fig2}a. With the increase of the distance, the variation of the uncertainty also increases, especially for the x and z components. However, the uncertainty at y direction is always smaller than the other two, especially at the distance greater than 40 meters from the sensor and its variation is also much smaller than those of the other 2 components. This might be explained by Fig. \ref{fig2}b which shows the negative correlation between the total uncertainty of the bounding box and IoU. It means that a more accurate prediction is usually accompanied by lower uncertainty. We further calculate the average corner loss of the $y$ component and find that it is only 0.056, which is much lower than $x$ (0.187) and $z$ (0.304) as expected. 

\subsubsection{Point cloud distribution representation}

To describe the spatial distribution of the point cloud and determine whether the corner is at the denser or the sparser side, we calculate the average Euler distance from each corner to all the detected points in the bounding box using (\ref{eq5}):
\begin{equation}
\label{eq5}
d_k = \frac{1}{N} \sum_{i} ||\mathbf{c}_k-\mathbf{p}_i||
\end{equation}
in which $d_k$ is the average Euler distance of $k$th corner of the bounding box,  $\mathbf{c}_k$ and $\mathbf{p}_i$ are the coordinates of the $k$th corner and $i$th detected point respectively. $N$ is the total number of the detected points within the bounding box.  

As to the uncertainty, we calculate the ensemble variance $\sigma_{ens}^2$  of each corner by summing up the variance of their own three components.

After the normalization in (\ref{normalize}), $d_k$ and $\sigma_{ens, k}$ can be regarded as the samples of two pseudo probabilistic distributions denoted as $D(k)$ and $U(k)$ respectively. To evaluate the similarity of the two distributions, i.e. how relevant are the distances and ensemble uncertainties in a corner set, we calculate the KLD of $U(k)$ to $D(k)$ as shown in (\ref{KLDUD}). Notice that $D(k)$ and $U(k)$ represent the proportional relationship of the distances and uncertainties of the eight corners respectively, and the KLD here denotes the information loss when $U(k)$ is used to approximate $D(k)$. Lower KLD means closer proportional relationship between the uncertainties and distances in a corner set. 

\begin{equation}
\label{normalize}
\begin{aligned}
d_k& \leftarrow d_k/	\rm{sum}(\mathbf{d}), \\
\sigma_{ens, k}& \leftarrow \sigma_{ens, k}/\rm{sum}(\mathbf{u})
\end{aligned}
\end{equation}

\begin{equation}
\label{KLDUD}
KLD = \sum_k D(k) \log{\frac{D(k)}{U(k)}}
\end{equation}

\begin{figure}[thpb]
  \centering
  \includegraphics[width=8cm]{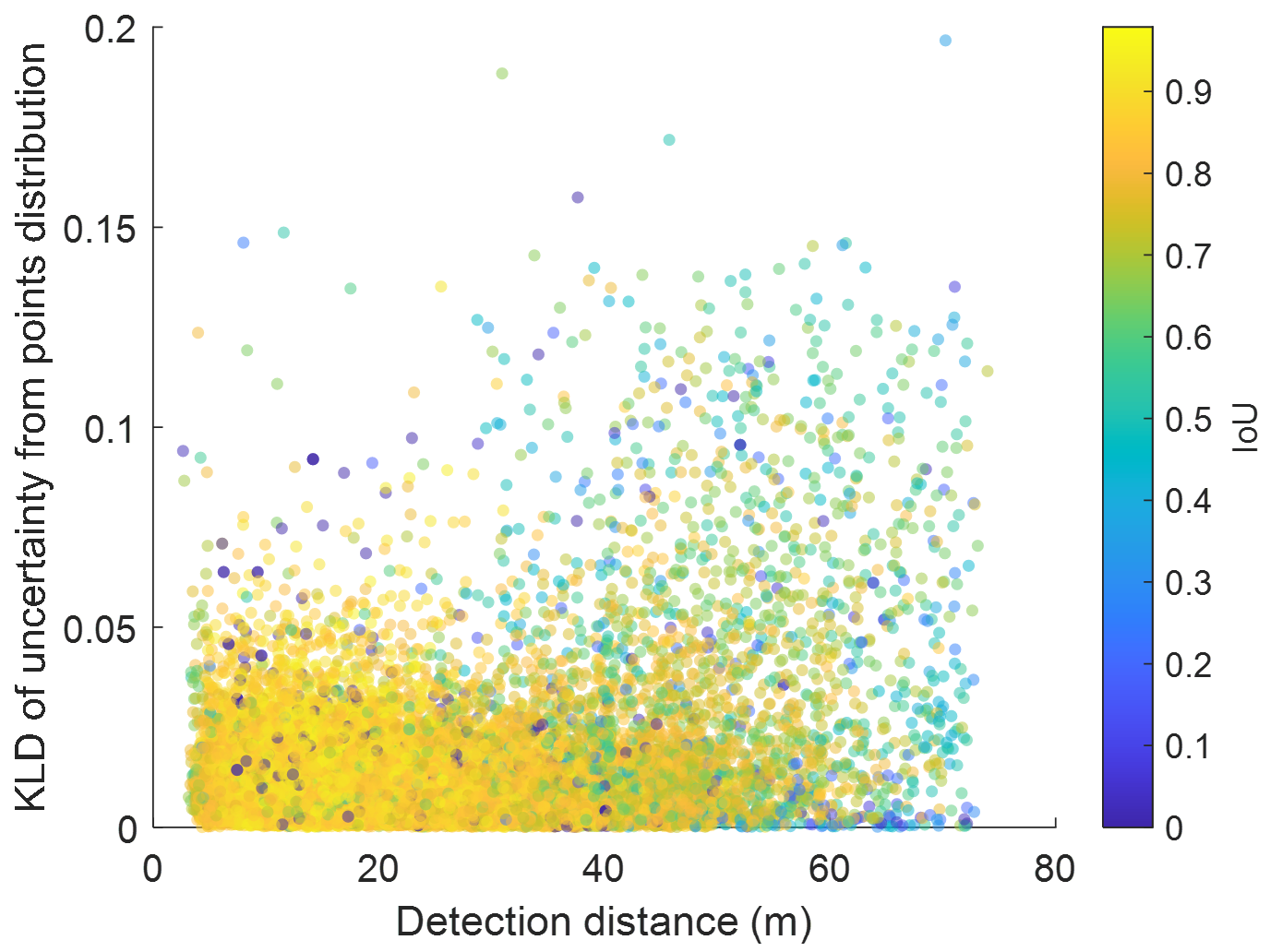}
  \caption{KLD of corner variance distribution from corner Eular distance distribution with respect to detection distance. The color represents the IoU of the bounding box}
  \label{KLDUDvsD}
\end{figure}

We plot the KLD with respect to the detection distance in Fig. \ref{KLDUDvsD} in which the overall KLD locates close to 0. With the increase of the detection distance, the number of the data points with KLD greater than 0.05 increases. After looking into the samples with high KLD, we find that most of these data points represent the bounding box with low IoUs which indicates they are less accurate predictions. 

\begin{figure}[thpb]
  \centering
  \includegraphics[width=8cm]{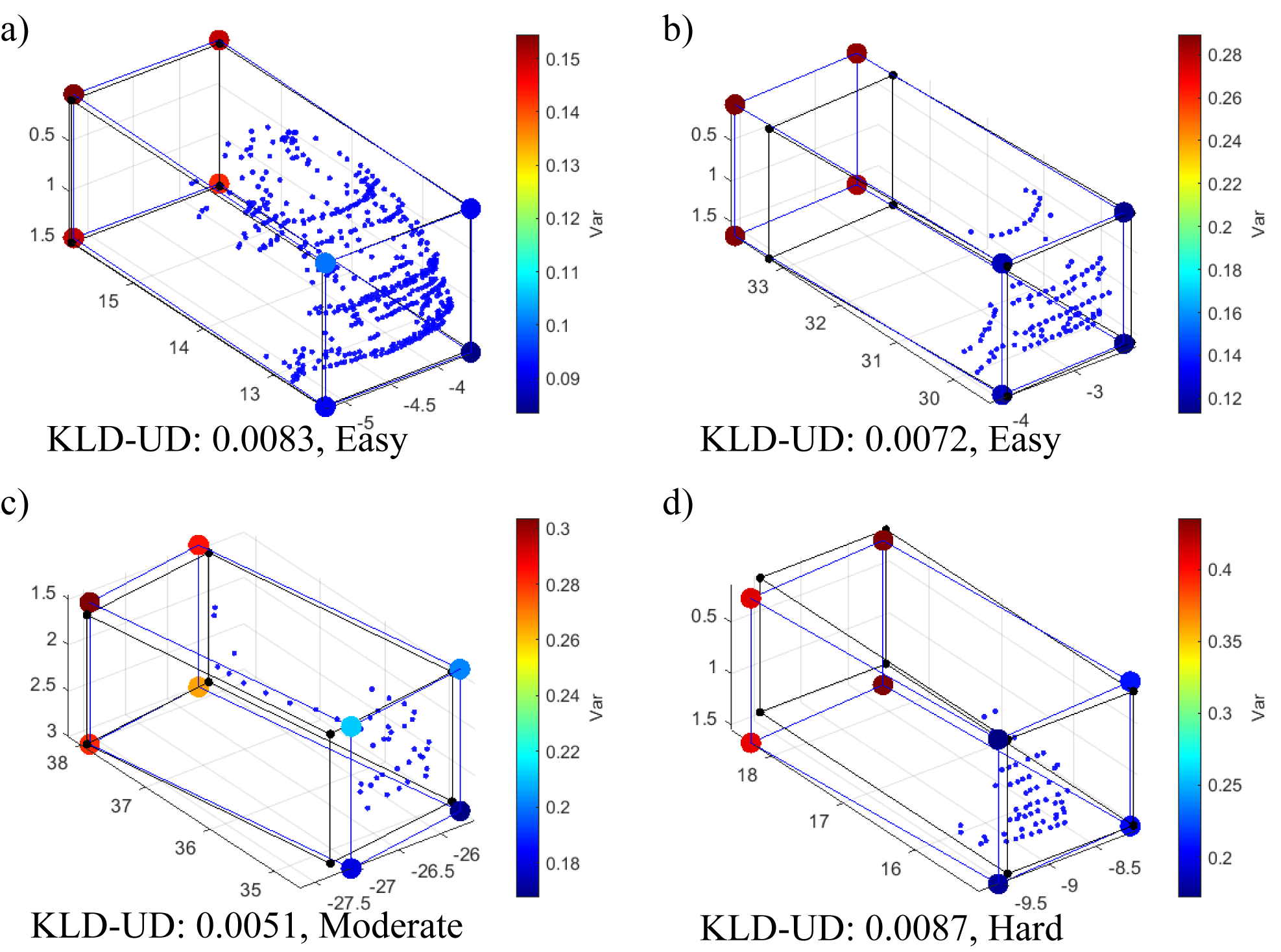}
  \caption{Sample analysis of the corner uncertainties at different difficulty levels. The ground truth boxes are with black lines and corners. The predicted boxes have blue lines and their corners are enlarged and filled with false color to represent the uncertainty from low (dark blue) to high (dark red). The detected points in the bounding box are plotted as blue points. For each subplot, we labeled them with their KLD of $U(k)$ from $D(k)$ (KLD-UD) and difficulty level.}
  \label{sampleUD}
\end{figure}

We further pick four samples and plot them with their KLD values within different difficulty levels in Fig. \ref{sampleUD} to analyze how they perform with low KLD values. As is defined in (\ref{KLDUD}), lower KLD literally means higher similarity of the distributions of the corner uncertainty and the point cloud. As indicated in Fig. \ref{sampleUD}, even at different difficulty levels and with different point numbers, the uncertainty of corner shows the same trend that it is lower at denser point cloud side. Fig. \ref{sampleUD}b shows that the predicted corners at the side with denser point cloud are closer to the ground truth than the spaser side with higher confidence. It matches our observation about the negtive correlation between the uncertainty and accuracy in the general behaviors of the uncertainty discussed in the former section. This is practical and would help to predict collisions in autonomous driving since the denser point side is most likely the closer side to the LiDAR. 

\begin{figure}[thpb]
  \centering
  \includegraphics[width=8cm]{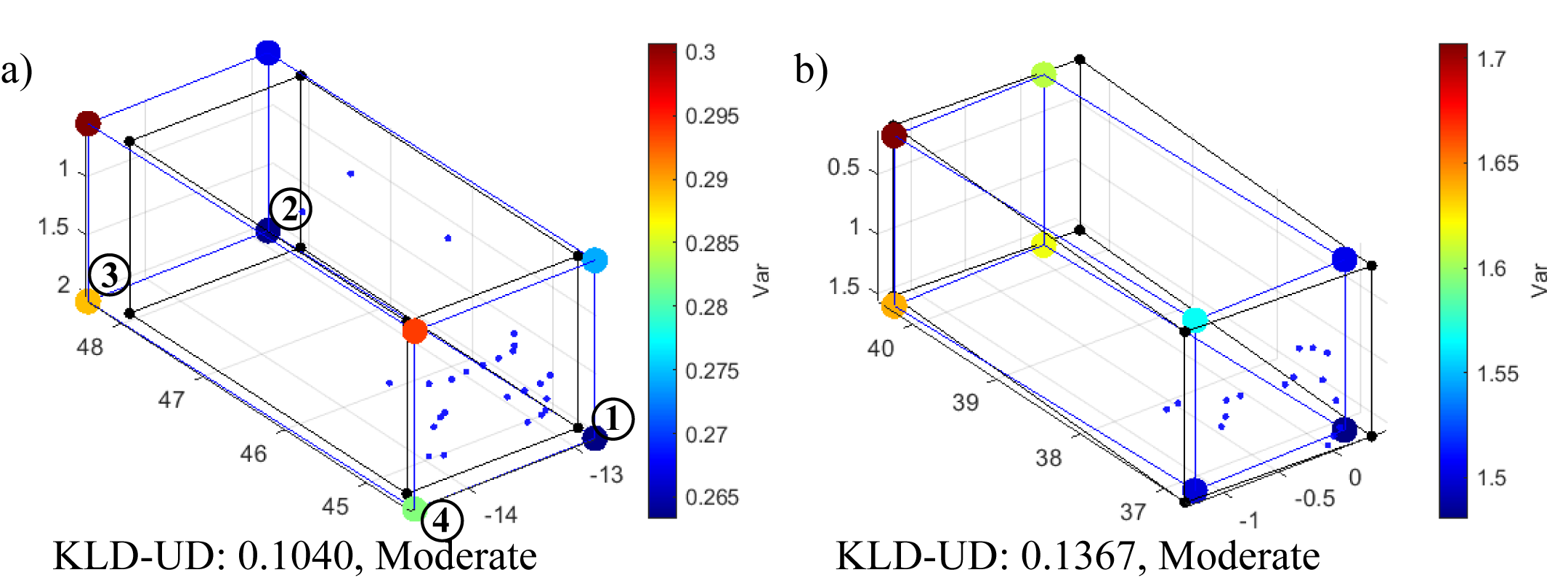}
  \caption{Samples with relatively higher KLD-UD. Feature discription of the figure can be refered in the caption of Fig. \ref{sampleUD}}
  \label{sampleUD_o}
\end{figure}

Besides most of the low KLD cases, we also pick two samples with relatively high KLD to analyze the ‘outliers’. While generally, as seen in Fig. \ref{sampleUD_o}, they are still the cases that we discussed in the former paragraph. In Fig. \ref{sampleUD_o}a, corners at the side with more points have lower uncertainties comparing to the other side which is also found in Fig. \ref{sampleUD_o}b. One reason to explain the higher KLD is that the uncertainty distribution does not exactly fit the point clouds corner by corner. For instance, in Fig. \ref{sampleUD_o}a, the uncertainty of corner 1 is smaller than that of corner 3 which agree on the point cloud distribution, while corner 2 and corner 4 are the opposite. Another reason of higher KLD is that these objects are of large  distance from the sensor which makes the total variance of the bounding box at a high level. Moreover, the differences between uncertainties of different corners is not as distinguishable as that in the point cloud distribution which makes the two distributions numerically not similar to each other. As seen in Fig. \ref{corntrans}b,  the variances of the corners vary from 1.5 to 1.7, whose change rate is only approximately 15\%, while those of the cases in Fig. \ref{sampleUD} show at least a 75\% difference. 

\subsubsection{Influence of geometry constraint}

Although we set a high DOF in modeling the uncertainties, the model still learns the constraint of the bounding box from its geometry. We use the relevant locations of corners to describe the geometry because they are determined by the parameters of the bounding box. For instance, once the shape of the box is set and we assume a point of the bounding box as the static reference, the relative displacement of the corners caused by the small pose variation are constrained by relative locations of the corners from the reference point. And the small displacement is approximately proportional to the distance between the corner to the reference point. This is also the way that small error transfers which can be represented using variance. 

In our case, we set the corner with the minimum variance as the reference point and its variance as the reference uncertainty. We calculate the variance difference and Euler distance between each corner and the reference point as shown in (\ref{KLDR})
\begin{equation}
\begin{aligned}
\label{KLDR}
\sigma_k& \leftarrow \sqrt{\sigma_{k}^2 - \rm{min}{\left\{\sigma_{i}^2\right\}} }, i \in \left\{ 1,2,...,7,8\right\} \\
d_{c,k}& \leftarrow ||\mathbf{c}_k - \mathbf{c}_m||
\end{aligned}
\end{equation}
in which  $m=\argmin\limits_i \left\{\sigma_i^2\right\}$, and $k$ is the index of the corner.

Like (\ref{normalize}) and (\ref{KLDUD}), we normalize $\sigma_k$ and $d_{c,k}$  with their sums respectively and calculate the KLD of the pseudo distribution $R_{\sigma}(k)=\sigma_k$ from the distribution $R_d(k)=d_{c,k}$. The KLD here represents the similarity between the distribution of relative predicted uncertainties $R_{\sigma}(k)$ and relative locations $R_d(k)$. Low KLD means that the predicted uncertainty tends to be constrained by the cuboid geometry of the bounding box based on a confident corner.  

\begin{figure}[thpb]
  \centering
  \includegraphics[width=8cm]{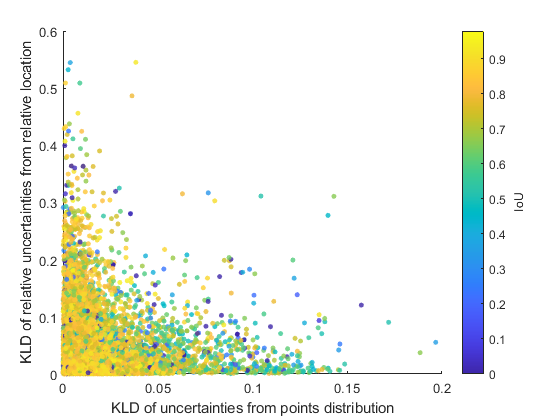}
  \caption{KLD of relative predicted uncertainties from relative locations with respect to KLD of corner variance distribution from corner Euler distance distribution, the color represents the IoU of the bounding box}
  \label{KLD-UD-R}
\end{figure}

We plot the KLD of relative predicted uncertainties from relative locations (KLD-R for short) with respect to the KLD of corner variance distribution from corner Euler distance distribution defined in (\ref{KLDUD}) (KLD-UD for short) in Fig. \ref{KLD-UD-R} to reveal the relationship between the influence of point cloud distribution and cuboid constraint on uncertainties. We find that most of the samples locate close to the axis which means the uncertainties agree on at least one of the two distributions $R_d(k)=d_{c,k}$ and $D(k)=d_k$. 

\begin{figure}[thpb]
  \centering
  \includegraphics[width=8cm]{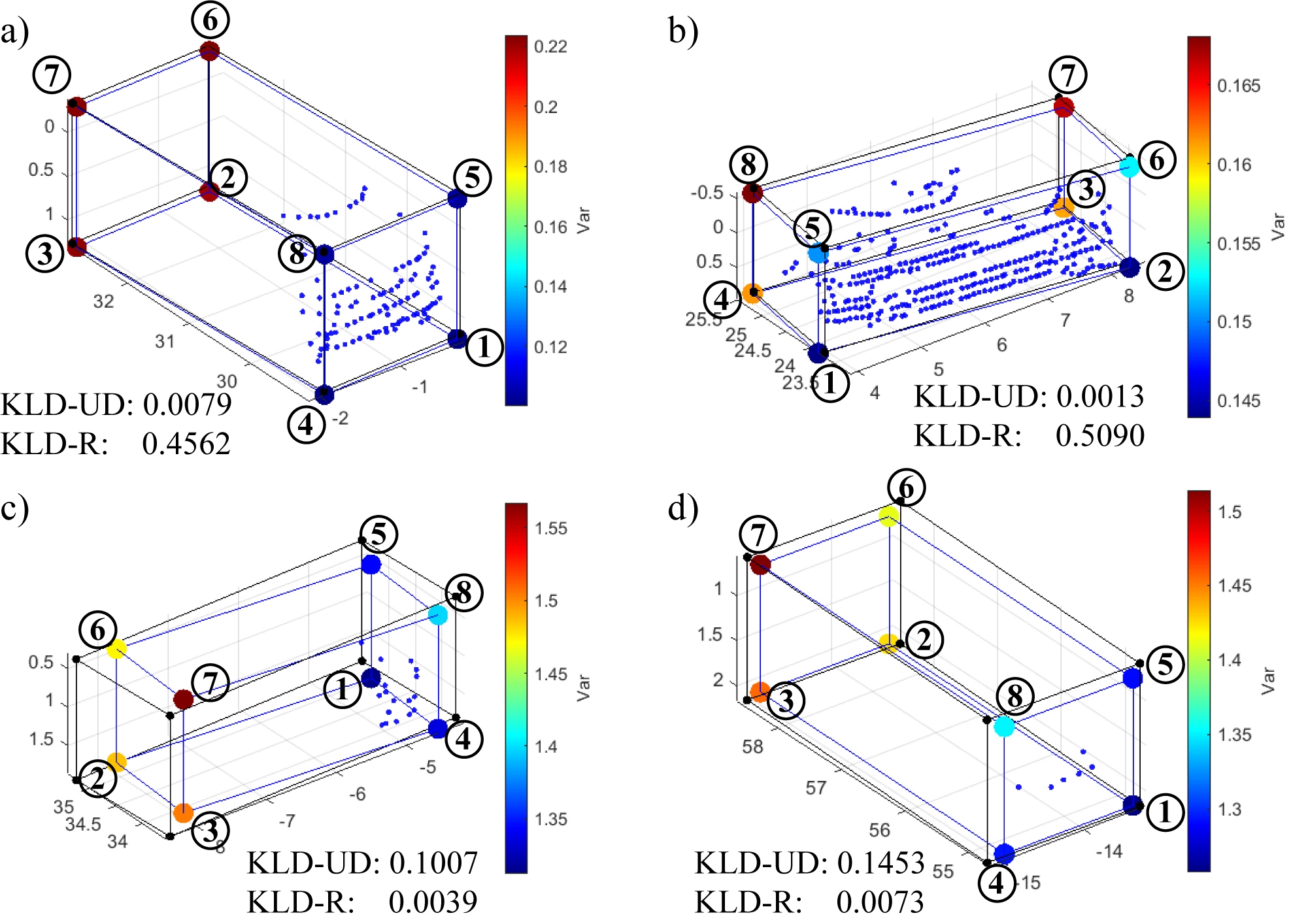}
  \caption{Representative sample with the different values of KLD-UD and KLD-R. a) and b) are with higher KLD-R and lower KLD-UD while c) and d) are the opposite. Feature discription of the figure can be refered in the caption of Fig. \ref{sampleUD}}
  \label{sampleKLDR}
\end{figure}

To better understand how the model learns the point cloud distribution and how it is affected by the cuboid constraints, we plot four representative samples in Fig. \ref{sampleKLDR}. As seen in Fig. \ref{sampleKLDR}a and \ref{sampleKLDR}b, the samples with higher KLD-R are with rich point cloud information, and the model predicts a confident face (formed by corner 1, 5, 8 and 4) in Fig. \ref{sampleKLDR}a while a confident edge (formed by corner 1 and 2) in Fig. \ref{sampleKLDR}b rather than a confident reference point in our designed test. Transferring our concept of relative point to the face and edge, we find the uncertainties of the samples in Fig. \ref{sampleKLDR}a and \ref{sampleKLDR}b are still constrained by the cuboid geometry. In Fig. \ref{sampleKLDR}b, denote the edge with corner $i$ and $j$ as “edge $ij$”, if we set edge 12 as the reference, edge 78 has the highest corner uncertainty which is also the farthest from the reference. While the other two edges are closer to the reference and have lower uncertainty comparing with edge 78. With less point information provided, the model is not able to predict a confident face or edge but only a confident point which results in lower KLD-R and potentially higher KLD-UD. As we can see in Fig. \ref{sampleKLDR}c and \ref{sampleKLDR}d, with limited point cloud information provided, the model tends to predict a relatively confident corner (corner 1 in both Fig. \ref{sampleKLDR}c and \ref{sampleKLDR}d), and the value of uncertainties of other corners are affected by their relative locations to the confident one. 

\section{Uncertainty Recovery}

In this section, we propose an efficient Bayesian updating method to recover the uncertainty of the original parameters of the bounding box. The basic idea is to divide the eight corners of the bounding box into 4 pairs. With proper division, we are able to derive the required parameters from each pair and denote the process as an individual measurement. Then, we calculate the variance of the obtained parameter using error transfer formula. Finally, we conduct Bayesian update to obtain the final uncertainties of the original parameters. 

\subsection{Uncertainty of an individual measurement}

To recover the yaw angle, we pick the edge that was parallel to the orientation of the car and utilized its two vertices as the corner pair. Denote the coordinates of the two corners in $x-z$ plane are $(x_i, z_i)$ and $(x_j, z_j)$. Then the yaw angle: 
\begin{equation}
    \label{yaw}
    \psi = \arctan \frac{z_i - z_j}{x_i - x_j}
\end{equation}

Applying the error transfer formula, we can obtain the variance of the yaw angle: 

\begin{equation}
\begin{aligned}
    \sigma_{\psi}^2 &= \left|\frac{\partial{\psi}}{\partial{x_i}}\right|^2\sigma_{xi}^2 + \left|\frac{\partial{\psi}}{\partial{x_j}}\right|^2\sigma_{xj}^2 + \left|\frac{\partial{\psi}}{\partial{z_i}}\right|^2\sigma_{zi}^2 + \left|\frac{\partial{\psi}}{\partial{z_j}}\right|^2\sigma_{zj}^2 \\
    &= \frac{|z_i-z_j|^2(\sigma_{xi}^2+\sigma_{xj}^2)+|x_i-x_j|^2(\sigma_{zi}^2+\sigma_{zj}^2)}{\left[(x_i-x_j)^2+(z_i-z_j)^2\right]^2}
\end{aligned}
\end{equation}

Similarly, we can obtain the uncertainties of the dimensions of the box  
\begin{equation}
    \sigma_{d, k}^2 = \frac{\sum_k (c_{i, k}-c_{j, k})^2(\sigma_{i, k}^2+\sigma_{j, k}^2)}{\sum_k (c_{i, k}-c_{j, k})^2}
\end{equation}
and the uncertainty of the location of the box 
\begin{equation}
    \sigma_{loc,k}^2=\frac{1}{2}(\sigma_{i, k}^2 + \sigma_{j, k}^2)
\end{equation}
in which $c_k, k\in \{1,2,3\}$ is the component of the corner. 

\subsection{Bayesian update}
With the variance obtained from the measurements discussed above, we can use Bayesian update method to approximate the final variance with (\ref{baysien}) \cite{lynch2007introduction}.

\begin{equation}
\label{baysien}
    \sigma_{bayesien}^2=\frac{\prod_i \sigma_i^2}{\sum_j \prod_{i \neq j} \sigma_i^2}
\end{equation}

\section{Discussions and Conclusions}
 We have presented an innovative design of loss function to improve the performance of a 3D object detector and learn an explainable uncertainty for the predictions. By applying corner transformation and uncertainty modeling, our method re-weights the original parameters of the bounding box in the loss function and increases the adaptivity of the model to the noisy and biased LiDAR data and labels. The test result on the KITTI \textit{val} split shows that the performance of our method increases by up to 15\% comparing with the baseline which is with simple L1 loss. As for the results on KITTI \textit{test} set, our method surpasses the original PointRCNN at moderate and hard difficulty level by 1.23\% and 2.47\% respectively which indicates better performance and higher robustness. 
 
To study the characteristics of the uncertainty, we design KLD-based tests to explain how the predicted uncertainties of the corners represent the distribution of the point cloud in the bounding box and how they are constrained by the cuboid geometry of the bounding box. As we expected, our method predicts lower uncertainties for corners at the side with relatively denser point cloud. Moreover, the distribution of the predicted uncertainties is constrained by the cuboid geometry of the bounding box in different cases based on the representation of the asymmetrically distributed point cloud is in the bounding box. 

With the method proposed in Section V, we can estimate the uncertainty of the parameters of the bounding box from the uncertainties of corners. What’s more, our method can be transferred to most deep-learning-based object detectors with little increment of computation cost. It not only increases the performance but also predicts convincing uncertainties for tracking and planning. We will further apply our method in the RPN stage for more improvement and test it on different state-of-the-art models to confirm its constancy.

\bibliographystyle{IEEEtran}
\bibliography{ref}

\end{document}